%% file: main.tex
\documentclass[runningheads]{llncs}
\usepackage{graphicx}
\usepackage{comment}
\usepackage{amsmath,amssymb}
\usepackage{color}

\usepackage{cite}
\usepackage{subfig}
\usepackage{multirow}
\usepackage{algpseudocode}
\usepackage{algorithm}

\usepackage[pagebackref=true,breaklinks=true,letterpaper=true,colorlinks,bookmarks=false]{hyperref}

\def\ourmethod{\textit{MIRAGE}}

\input{cmd.tex}

\begin{document}
\pagestyle{headings}
\mainmatter
\def\ECCVSubNumber{3452}

\title{Improving Deep Metric Learning with Virtual Classes and Examples Mining}

\author{Pierre Jacob$^1$ \and David Picard$^2$ \and Aymeric Histace$^1$ \and Edouard Klein$^3$}
\authorrunning{P. Jacob et al.}
%
\institute{ETIS UMR 8051, Universit{\'e} Paris Seine, UCP, ENSEA, CNRS, F-95000, Cergy, France\\
\email{\{pierre.jacob, aymeric.histace\}@ensea.fr}\and
LIGM, UMR 8049, \'Ecole des Ponts, UPE, Champs-sur-Marne, France \email{david.picard@enpc.fr} \and
C3N, P\^{o}le Judiciaire de la Gendarmerie Nationale, 5 boulevard de l'Hautil, 95000 Cergy, France\\
}
\maketitle

\begin{abstract}
    In deep metric learning, the training procedure relies on sampling informative tuples.
    However, as the training procedure progresses, it becomes nearly impossible to sample relevant hard negative examples without proper mining strategies or generation-based methods.
    Recent work on hard negative generation have shown great promises to solve the mining problem.
    However, this generation process is difficult to tune and often leads to incorrectly labeled examples.
    To tackle this issue, we introduce \ourmethod, a generation-based method that relies on virtual classes entirely composed of generated examples that act as buffer areas between the training classes.
    We empirically show that virtual classes significantly improve the results on popular datasets (Cub-200-2011, Cars-196 and Stanford Online Products) compared to other generation methods.
\end{abstract}

\input{tex/intro.tex}

\input{tex/rw.tex}

\input{tex/method.tex}

\input{tex/results.tex}
\input{tex/ccl.tex}

\bibliographystyle{splncs04.bst}
\bibliography{biblio}

\end{document}

%% file: cmd.tex
\usepackage{amsmath,amsfonts,bm}

\def\eqref#1{equation~\ref{#1}}

\def\1{\bm{1}}

\newcommand{\test}{\mathcal{D_{\mathrm{test}}}}

\def\vf{{\bm{f}}}

\def\vp{{\bm{p}}}

\def\vx{{\bm{x}}}

\def\mM{{\bm{M}}}

\def\mW{{\bm{W}}}

\DeclareMathAlphabet{\mathsfit}{\encodingdefault}{\sfdefault}{m}{sl}
\SetMathAlphabet{\mathsfit}{bold}{\encodingdefault}{\sfdefault}{bx}{n}



%% file: tex/intro.tex
\section{Introduction}\label{sec:intro}
    \begin{figure}[t]
        \centering
        \hfill
        \subfloat[Standard hard negative generation\label{fig:illustration_failure}] {\includegraphics[width=0.3\columnwidth]{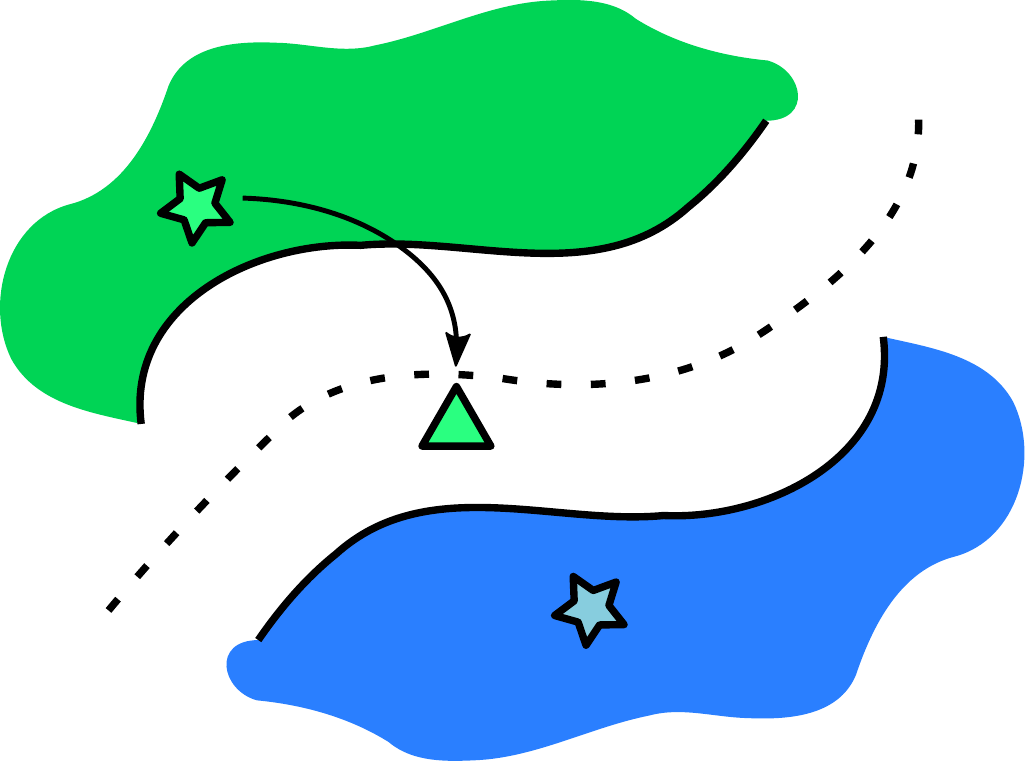}}
        \hfill
        \subfloat[Virtual class (dashed line manifold) hard negative generation\label{fig:illustration_solved}] {\includegraphics[width=0.35\columnwidth]{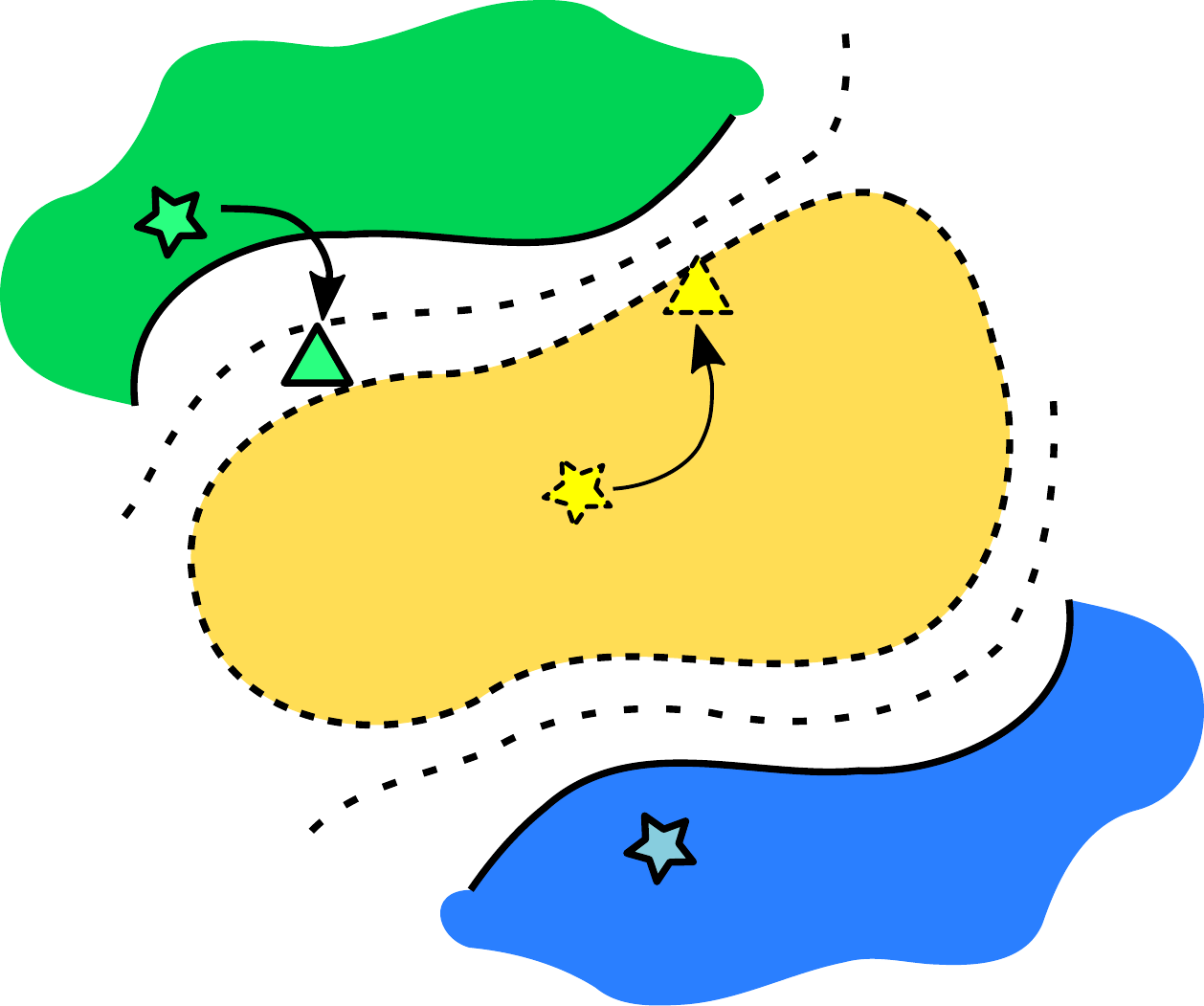}}
        \hfill
        \hfill
        \caption{Hard negative generation. The standard hard negative generation on \autoref{fig:illustration_failure} can lead to incorrect label if the generated example is sampled beyond the boundary of the class manifold.
        By adding a virtual class between the training classes on \autoref{fig:illustration_solved}, hard negative examples generated beyond the boundary of the class manifold are still within the correct classes with respect to training classes.}
        \label{fig:illustration_method}
    \end{figure}
    
    Deep metric learning (DML) is an important, yet challenging task in the Computer Vision community, with numerous applications such as multi-modal retrieval \cite{Carvalho_2018_SIGIR, Wehrmann_2018_CVPR}, face verification \cite{Schroff_2015_CVPR} or person re-identification \cite{Liu_2016_CVPR}.
    DML methods intend to learn an embedding space, where visually-related images (\emph{e.g.}, two different birds from the same breed) have similar representations, while unrelated images (\emph{e.g.}, two different breeds of crows from North America and Europe) have dissimilar representations.
    To learn this embedding space, recent contributions focus on three main points: (1) loss functions to improve generalization \cite{Wang_2019_CVPR}, (2) ensemble methods to tackle the embedding space diversity \cite{Opitz_2017_ICCV} and (3) hard example mining strategies to resume the training when randomly sampling informative tuples becomes nearly impossible \cite{Xu_2019_CVPR}.
    
    Example generation has recently been proposed as a hard negative mining strategy.
    In this case, a generator and the metric learning network are trained together to provide informative tuples using either VAEs \cite{Lin_2018_ECCV} or GANs \cite{Duan_2018_CVPR, Zhao_2018_ECCV, Zheng_2019_CVPR}.
    In the case of VAEs, a large amount of examples is generated by sampling with respect to the training sample distribution estimated from the data.
    Usually, this leads to sampling inside the class manifolds and rarely produces hard negative examples.
    Such variational approaches are interesting in the case of few training samples per class but they are not well suited for mining informative examples at later training stages.
    On the opposite, GAN-based approaches generate discriminative examples.
    However, adversarial generators are difficult to tune due to the contrary objectives of the DML network and the adversarial learning of the generator.
    On the one hand, if the adversarial loss is much lower than the DML loss, the generated examples tend to be at the center of the class manifold and the method faces the same problems as VAE generators.
    On the other hand, if the adversarial loss is much higher than the DML loss, some examples can be generated beyond the boundary of the class manifolds and lead to label ambiguity as illustrated on \autoref{fig:illustration_failure}.
    The mining strategy then produces examples with incorrect labels with respect to the training classes.
    
    As the main contribution of this paper, we propose \ourmethod , a method that leverages virtual classes composed solely of generated examples to tackle the problem of label ambiguity arising from hard negative generation.
    Virtual classes play the role of buffer areas as shown on \autoref{fig:illustration_solved}.
    Hard negative examples that lie between the training class manifolds are generated inside these buffer areas, without any label ambiguity, by sampling the virtual classes.
    In addition to solving the problem of label ambiguity, virtual classes example generation leads to better generalization capabilities: The metric learning network has better results on unseen classes than other adversarial approaches \cite{Duan_2018_CVPR, Zhao_2018_ECCV, Zheng_2019_CVPR}.

    
    The paper is organized as follow: in \autoref{sec:rw}, we present the related work in deep metric learning, the recent contributions in example generation and motivate the need for our method.
    In \autoref{sec:met}, we expose the core aspects of \ourmethod \ and its simple implementation.
    In section \autoref{sec:abla}, we experimentally show that \ourmethod \ indeed produces buffer areas between the training classes and we perform an ablation study of the different aspects of the method.
    Finally, in \autoref{sec:exp}, we show that our method improves over other sample generation methods on four DML datasets (Cub-200-2011, Cars-196, Stanford Online Products and In-Shop Clothes Retrieval), and obtains results comparable to the state-of-the-art.

%% file: tex/rw.tex
\section{Related work}\label{sec:rw}
    In deep metric learning, we train a deep network to provide representations and a corresponding metric to measure similarities.
    The training procedure of such network relies on three main points, namely: a loss function, a sampling strategy and optionally, an ensemble method.
    In the case of the loss function, original methods consider pairs \cite{Chopra_CVPR_2005} or triplets \cite{Schroff_2015_CVPR} of similar/dissimilar examples.
    These approaches have been enhanced by considering larger tuples \cite{Chen_2017_CVPR, Song_2016_CVPR, Sohn_2016_NIPS, Ustinova_2016_NIPS} or by improving the properties of the loss functions \cite{Rippel_2016_ICLR, Wang_2018_CVPR, Wang_2017_ICCV, Yu_2018_ECCV}.
    When randomly sampling informative tuples becomes too hard, sampling strategies can be exploited to resume the training.
    These methods can be based on efficient batch construction \cite{Song_2016_CVPR, Sohn_2016_NIPS, Ustinova_2016_NIPS}, scalable mining strategies \cite{Harwood_2017_ICCV, Schroff_2015_CVPR} or proxy-based approximations \cite{Movshovitz-Attias_2017_ICCV, Song_2017_CVPR, Rippel_2016_ICLR}.
    Finally, ensemble methods have become a popular way of improving the performances of DML architectures \cite{Kim_2018_ECCV, Opitz_2017_ICCV, Opitz_toap_PAMI, Xuan_2018_ECCV, Yuan_2017_ICCV}.
    Our proposed \ourmethod \ is a complementary approach to loss functions and ensemble methods.
    
    \ourmethod \ differs from other hard negative example generation methods, such as DAML \cite{Duan_2018_CVPR}, HTG \cite{Zhao_2018_ECCV}, HDML \cite{Zheng_2019_CVPR} and DVML \cite{Lin_2018_ECCV}.
    Both DAML and HTG rely on sampling a triplet, by feeding this triplet to a generator trained in an adversarial manner, and then producing a hard negative example to replace the original negative one.
    However, both methods suffer from the problem of label ambiguity illustrated in \autoref{fig:illustration_method} in that the generator can output an example inside another class manifold.
    Zheng \emph{et al.} \cite{Zheng_2019_CVPR} face the same issue with HDML.
    HDML tries to alleviate this effect by generating first an intermediate example that may be outside its class manifold; then a generator projects this example into the class manifold.
    In case of a failure, they use the DML loss over the real examples to weight the generator loss: if the triplet is an easy one, the generator only slightly modifies the example to avoid the generation of an intermediate example that would be too far from its class manifold.
    Moreover, the metric learning loss that is computed on the generated triplets is also weighted by the reconstruction loss: the worse the reconstruction is, the less they take into account the new triplet.
    In other words, to mitigate the effect of label ambiguity, HDML tends to discard really hard negative examples which limits its hard negative generation capabilities.
    
    At the same time, DVML gets rid of the triplet constraints for the generator by considering the class manifold as a Gaussian distribution.
    By estimating the parameters of the distribution, the sampling of new examples is performed using a variational approach.
    Because there is no adversarial training, examples tend to be mostly sampled at the center of the Gaussian distribution.
    As such, they only slightly contribute to the DML loss.
    
    To solve the problem of label ambiguity while generating hard negative, we propose to insert buffer areas between the training classes.
    To that end, we introduce virtual classes that we encourage to migrate between the training classes.
    Sampling hard negatives from the virtual classes allows us to use a generative sampling process similar to DVML \cite{Lin_2018_ECCV} which is simpler than the triplet based adversarial methods.
    At the same time, it also removes the need to take into account the possible incorrectness of the labels since these generated examples do not correspond to existing classes. 

%% file: tex/method.tex
\section{Method overview}\label{sec:met}
    In this section, we start by giving an overview of \ourmethod.
    Then, we detail the core aspects involved in the approach.
    Finally, we describe the overall \ourmethod \ architecture.
    
    \subsection{\ourmethod \ overview}\label{sec:met_overview}
        \ourmethod \ is designed to improve deep metric learning by using the following core aspects:
        
        \textbf{DML training.}
            Like any other DML method, \ourmethod \ uses a deep neural network to embed feature vectors into a latent representation space where visually-related images have similar representations and where unrelated images have dissimilar representations.
            We use the standard metric learning approach which extracts deep local features using a backbone network (\emph{e.g.}, GoogleNet \cite{Szegedy_2015_CVPR} or BN-Inception \cite{Ioffe_2015_ICML}), computes a feature vector (\emph{e.g.}, using an average pooling) and projects it into an embedding space in order to learn the metric.
            
        \textbf{Training class sample generation.}
            As it is done in variational approaches, \ourmethod \ generates artificial examples from the training classes in order to provide a better sampling of each training class manifold.
            By doing so, class manifolds are filled with synthetic examples.
            These generated examples are added to the mini-batch along real examples in order to have larger batches from which informative tuples can be sampled.
            These additional tuples are then used to train the DML model.
            
        \textbf{Virtual class hard negative sample generation.}
            Similarly to adversarial sample generation, \ourmethod \ also generates hard negative examples.
            However, current hard negative generation are prone to label ambiguity (see \autoref{fig:illustration_failure}).
            To tackle this issue, we add virtual classes between training classes that play the role of buffer areas (see \autoref{fig:illustration_solved}).
            Hard negative examples are consequently generated inside these buffer areas by sampling within these virtual class manifolds.
            Similarly to training class generation, these examples are added to the mini-batch.
            We experimentally show that it leads to better performances than other generation-based methods.
        
    \subsection{Deep metric learning}\label{sec:met_dml}
        The first part of a DML network is to extract a feature vector $\vf_t$.
        \emph{E.g.}, we use GoogleNet \cite{Szegedy_2015_CVPR} followed by a global average pooling to compute $\vf_t$.
        Then, we want to learn a Mahalanobis distance $d_\mM$ so that the distance between two feature vectors $\vf_{t_i}$ and $\vf_{t_j}$ is:
        \begin{align}
            \nonumber d^2_\mM(\vf_{t_i}, \vf_{t_j}) &= \left( \vf_{t_i} - \vf_{t_j} \right)^\top \mM \left( \vf_{t_i} - \vf_{t_j} \right) \\
            \nonumber &= \left( \vf_{t_i} - \vf_{t_j} \right)^\top \mW \mW^\top \left( \vf_{t_i} - \vf_{t_j} \right) \\
            &= \|\mW^\top \vf_{t_i} - \mW^\top \vf_{t_j} \|_2^2
        \end{align}
        where $\mM = \mW \mW^\top$ is a low-rank approximation of $\mM$.
        The feature vectors are projected into the embedding space with $\mW$ where their corresponding examples are denoted $\vx_t$.
        In practice, all examples $\vx_t$ are $\ell_2$-normalized to ease the optimization.
        We note $E$ the function that transforms a feature vector $\vf_t$ into an example $\vx_t$ using the following equation:
        \begin{align}
            \vx_t = E(\vf_t) = \frac{\mW^\top\vf_t}{\|\mW^\top\vf_t\|_2}
        \end{align}
        
        To train the network, we rely on standard metric learning loss functions such as the contrastive loss \cite{Chopra_CVPR_2005}, the triplet loss \cite{Schroff_2015_CVPR} or the binomial loss \cite{Ustinova_2016_NIPS}.
        As proposed by \cite{Movshovitz-Attias_2017_ICCV}, we use a class representation prototype $\vp_t$ to accelerate the training. $E$ and $\vp_t$ are trained together using a DML loss (triplet, contrastive, \emph{etc.}) denoted $\mathcal{L}_\text{DML}$. 
        
    \subsection{Training class example generation}\label{sec:met_tc}
        To generate examples from the training classes, \ourmethod \ relies on a conditional generator $G$ that is designed to produce an artificial example $\widetilde{\vx}_t$ from the prototype $\vp_t$ of class $t$ and a Gaussian noise $\boldsymbol{\epsilon} \sim \mathcal{N}(\textbf{0}; \boldsymbol{\Sigma})$, as follows:
        \begin{align}\label{eq:training_class_generation}
            \widetilde{\vx}_t = E\left(G\left(\frac{\vp_t + \boldsymbol{\epsilon}}{\|\vp_t + \boldsymbol{\epsilon}\|_2}\right)\right)
        \end{align}
        $\widetilde{\vx}_t$ is then used as a training example to optimize the loss function described in the previous section.
        
        To train the generator $G$, we use a reconstruction loss by computing the ElasticNet loss between a feature vector $\vf_t$ extracted from a real image and a feature vector generated by feeding the generator G with $E(\vf_t)$, as follows:
        \begin{align}
            \mathcal{L}_{\text{rec}} = \|\vf_t - G(E(\vf_t))\|_1 + \|\vf_t - G(E(\vf_t))\|_2^2
        \end{align}
        
    \subsection{Virtual class example generation}\label{sec:met_vc}
        To generate hard negative examples, we consider a set of virtual classes associated with prototypes $\vp_v$.
        Examples are generated from these prototypes exactly like if they where training classes.
        To produce hard negative samples, we encourage the prototypes and the generator to output realistic samples between the training classes.
        To that end, we use a discriminative classifier $D$.
        $D$ is trained using binary cross-entropy to distinguish between real and generated samples (with output $D_{g}$).
        $D$ is also trained using categorical cross-entropy to predict classes (with output $D_c$).
        The combined loss $\mathcal{L}_\text{adv}$ for training $D$ is a two head classification loss based on cross-entropy:
        
        \begin{align*}
            \mathcal{L}_\text{adv} =& \underbrace{-\log(D_{g}(E(\vf_t))) -\log\left(1-D_{g}\left(E\left(G\left(\frac{\vp_v + \boldsymbol{\epsilon}}{\|\vp_v + \boldsymbol{\epsilon}\|_2}\right)\right)\right)\right)}_{\text{binary cross-entropy on real/generated samples}} \\
            & \underbrace{-y_t\log(D_{c}(E(\vf_t))) -y_v\log\left(D_{c}\left(E\left(G\left(\frac{\vp_v + \boldsymbol{\epsilon}}{\|\vp_v + \boldsymbol{\epsilon}\|_2}\right)\right)\right)\right)}_{\text{categorical cross-entropy on the classes}}
        \end{align*}
        where $y_t$ and $y_v$ are the class labels of the prototypes $\vp_t$ and $\vp_v$ respectively.
        
        To encourage the generator to output realistic samples that are between the training classes, $G$ is trained to maximize $\mathcal{L}_\text{adv}$, with $D$ being fixed.
        By optimizing over $D_g$, the generator is encouraged to output generated samples that are indistinguishable from real samples.
        By optimizing over $D_c$, the generator is encouraged to output samples at the boundaries of the classes (\textit{i.e.}, in the buffer area described in \autoref{fig:illustration_solved}).
        Just like the training class sample generation, virtual class sample generation is used to populate the mini-batches used for training using $\mathcal{L}_\text{DML}$.

    \subsection{\ourmethod \ architecture}\label{sec:met_arch}
    
\begin{figure}[t]
  \centering
  \includegraphics[width=\linewidth]{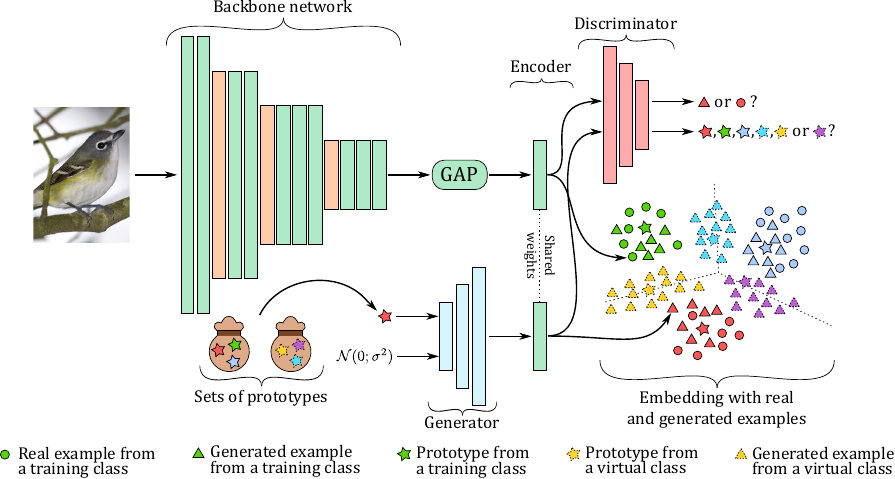}
  \caption{Overview of our proposed architecture. We extract a feature vector from the image using the backbone network and a global average pooling (GAP). Then, this feature vector is projected into the embedding space where we learn the metric. The framework relies on a set of prototypes (stars) which belong either to the training classes (plain lines) or to the virtual classes (dashed lines). We send through a generator a sampled prototype and a Gaussian noise to generate a new feature vector that is then projected into the same embedding space as the training images. In order to train this generator, we use a conditional discriminator to determine whether the sample is real or generated but also to determine the class to which it belongs.}
  \label{fig:archi_overview}
\end{figure}

        We describe the implementation of the \ourmethod \ architecture in \autoref{fig:archi_overview}.
        A set of deep local features is first extracted from the image using a backbone network such as GoogleNet \cite{Szegedy_2015_CVPR} or BN-Inception \cite{Ioffe_2015_ICML}.
        These local features are then aggregated into feature vectors using an average pooling.
        They are followed by the encoder $E$ which is composed of a single fully-connected layer without bias followed by a $\ell_2$ normalization.
        A prototype $\vp$ is used for each class and is represented by a star, either in plain lines for training classes or in dashed lines for virtual classes.
        The generator $G$ is composed of two fully-connected layers with ReLU activation.
        The discriminator $D$ is composed of a fully-connected layer with ReLU activation which is followed by two fully-connected layers: One with sigmoid activation for the binary classification of real or virtual feature vectors and one with softmax activation for the class prediction.
        
        To train \ourmethod, we generate mini-batches composed of training examples $\vx_t$, generated examples from the training class $\widetilde{\vx}_t$ and generated examples from virtual classes $\widetilde{\vx}_v$.
        The ratio of training examples and generated examples in the mini-batch corresponds to how much each aspect of \ourmethod \ is used. This ratio is investigated in the ablation studies.
        The backbone network, the encoder $E$ and the prototypes are trained together using the entire mini-batch minimizing the metric learning loss function $\mathcal{L}_\text{DML}$ from \autoref{sec:met_dml}.
        The generator $G$ is trained on $\widetilde{\vx}_t$ minimizing $\mathcal{L}_\text{rec}$ from \autoref{sec:met_tc} and on $\widetilde{\vx}_v$ maximizing $\mathcal{L}_\text{adv}$ from \autoref{sec:met_vc}.
        Finally, the discriminator $D$ is trained on the entire batch minimizing $\mathcal{L}_\text{adv}$ from \autoref{sec:met_vc}.

%% file: tex/results.tex
\section{Ablation study}\label{sec:abla}
    In this section, we provide different ablation studies that include (1) a visualization of the learned embedding and the virtual classes as well as some relevant statistics, (2) the impact of the number of generated example in the mini-batches, (3) the impact of the virtual class generation and (3) the combination of training class generation and virtual class generation.
    In this section, we train a GoogleNet backbone network with a 512 dimensional embedding with the contrastive loss on the Cub-200-2011 dataset that is denoted as the \emph{Baseline} using a fixed batch size of real examples $|\mathcal{B}|=40$ for all experiments.
    
    \subsection{Prototype visualization}
\begin{figure}
    \centering
    \subfloat[Cub-200-2011\label{fig:abla_cub}] {\includegraphics[width=0.47\linewidth]{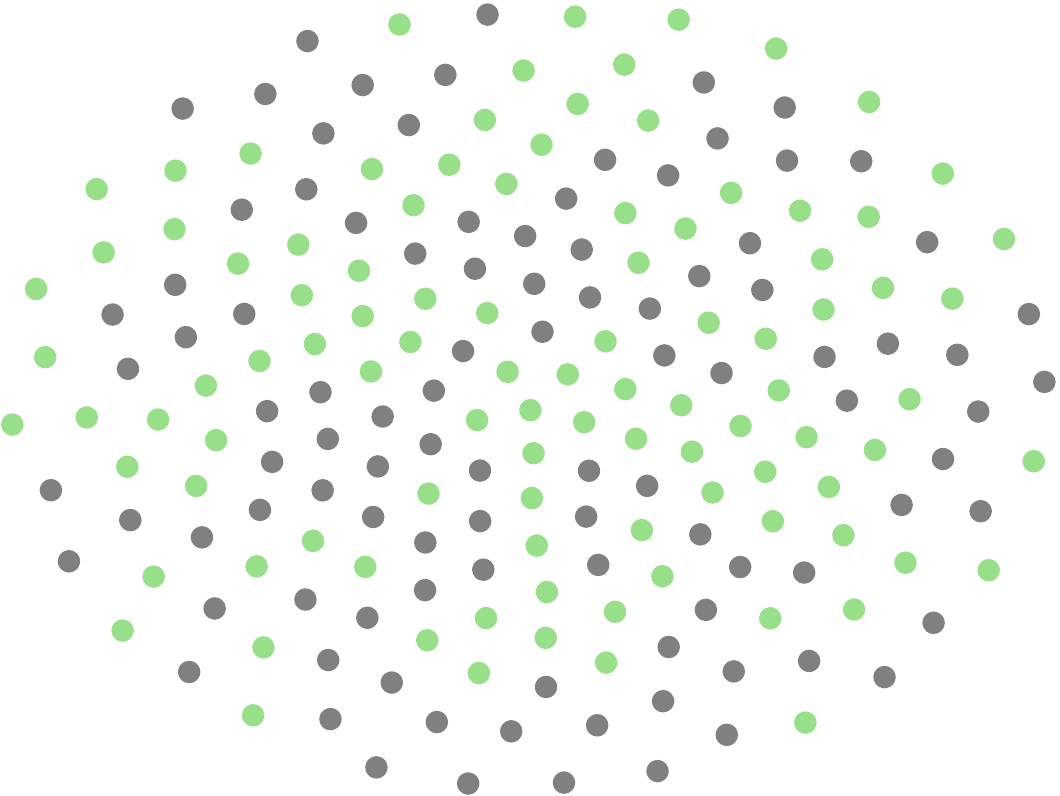}}\hfill
    \subfloat[MNIST\label{fig:abla_mnist}] {\includegraphics[width=0.47\linewidth]{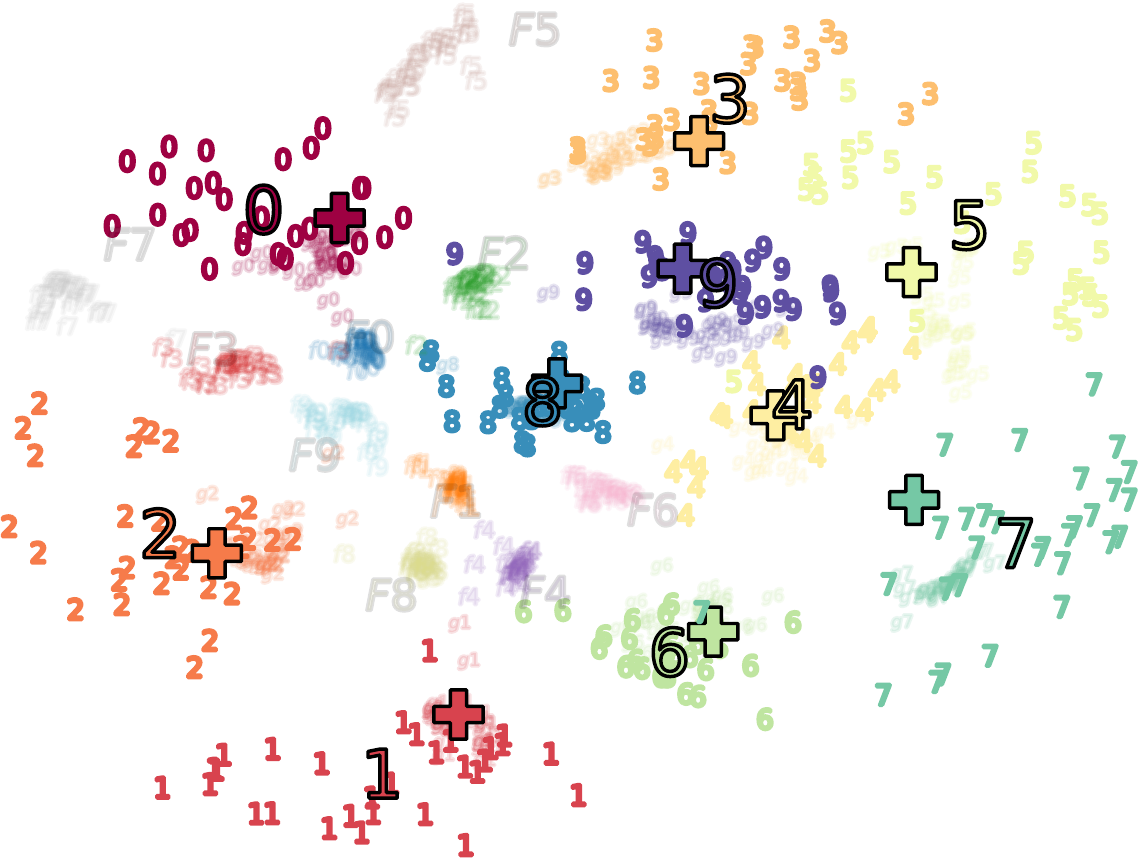}}
    \caption{Visualization of the training and virtual prototypes on MNIST and Cub-200-2011 datasets. For the Cub-200-2011 dataset (left), a t-SNE on the prototypes is run to visualize the high-dimensional embedding space. Virtual classes are in gray while training classes are in green. For the MNIST dataset (right), a 2D embedding space is learned without the $\ell_2$-norm. Training examples are represented by the numbers, and the prototypes are represented by the crosses. Virtual examples are in pale color.}
    \label{fig:my_label}
\end{figure}
        The first ablation considers an empirical analysis of the learned embedding space with the virtual classes.
        The objective is to verify that our architecture encourages the virtual classes to settle between the training classes.
        To that end, we show a t-SNE visualization of the training and virtual prototypes of the model trained on Cub-200-2011 on \autoref{fig:abla_cub}.
        As we can see, the virtual prototypes (in gray) are indeed in the middle of the training class prototypes.
        Quantitatively, we found that 80\% of the training class prototypes have a virtual class prototype as nearest neighbor in the 512 dimensional latent space.
        This numerically shows that our architecture is able to produce virtual class as buffer areas between training classes.
        
        To avoid the bias introduced by the 2D embedding performed by t-SNE, we also train a model on the popular MNIST dataset with a latent space of dimension 2.
        We plot the resulting prototypes as well as examples generated from these prototypes on \autoref{fig:abla_mnist}.
        As we can see, the virtual prototypes (denoted $F$ and in pale colors) are indeed acting as buffer between the training classes, even with the high constraints of having such a low dimensional latent space.
        
    \subsection{Sample generation ablation}
\setlength{\tabcolsep}{2.8pt}
\begin{table}
    \centering
    \begin{tabular}{|c|c|c|c|c|c|c||c|c|c|c|c|c|}\hline
         & \multicolumn{6}{|c||}{Training class examples ratio} &  \multicolumn{6}{c|}{Virtual class ratio} \\\hline
        ratio & 0\% & 10\% & 50\% & 100\% & 200\% & 400\% & 0\% & 10\% & 50\% & 100\% & 200\% & 400\% \\\hline
        Recall@1 & 57.0 & 57.8 & \textbf{58.8} & 58.5 & 58.7 & 58.7 & 57.0 & 58.2 & 59.0 & \textbf{59.3} & 58.6 & 58.9\\\hline
    \end{tabular}
    \caption{Impact of the number of training class generated examples and of the number of virtual class in the mini-batches. A ratio of 0\% means that no examples are generated.}
    \label{tab:vc_tc_ablation}
\end{table}
        First, we evaluate the impact of the number of generated training class examples.
        For that purpose, we do not use virtual class prototypes.
        We vary the size of the generated example set $\widetilde{\mathcal{B}}$ with respect to a ratio $r$ of the real example set $\mathcal{B}$, such that: $|\widetilde{\mathcal{B}}| = r \ |\mathcal{B}|$.
        We report Recall@1 on the Cub-200-2011 dataset in \autoref{tab:vc_tc_ablation} for $r \in \{0, 10\%, 50\%, 100\%, 200\%, 400\%\}$.
        
        The reported value for $r=0\%$ means that no examples have been generated and obtains a strong \emph{Baseline} of $57.0\%$ Recall@1.
        One can note that even a small amount of generated example, \emph{e.g.}, $10\%$, increases the performances by $0.8\%$ in Recall@1 on the Cub-200-2011 dataset.
        Hence, this confirms the benefit of a generation-based mining strategy to improve DML.
        With a further increase of the size of the generated example set, we improve the performances of the \emph{Baseline} from $57.0\%$ to $58.8\%$ Recall@1, a significant increase of nearly $2\%$.
        
        Next, we evaluate the impact of the number of virtual classes.
        We fix the size of the generated examples set $\widetilde{\mathcal{B}}$ to the size of the training examples batch $\mathcal{B}$, that is: $|\widetilde{\mathcal{B}}| = \ |\mathcal{B}| = 40$.
        Then, we vary the number of the virtual class prototype $N_v$ as a ratio of the number of the training class $N_t$, such that $N_v = r \ N_t$.
        We only generate examples from these virtual classes and not from the training classes.
        We report Recall@1 on the Cub-200-2011 dataset in \autoref{tab:vc_tc_ablation} for $r \in \{0, 10\%, 50\%, 100\%, 200\%, 400\%\}$.
        Interestingly, even a small number of additional classes, \emph{e.g.} $10\%$, already improves the \emph{Baseline} by a significant increase of more than $1.0\%$ in Recall@1, from $57.0\%$ to $58.2\%$.
        Increasing the number of virtual classes improves even more the performances, and leads to the best results for Recall@1 with $59.3\%$ - a significant increase of more than $2\%$ over the \emph{Baseline}.
        
        Finally, we evaluate the merging of both the training class example generation and the virtual class example generation.
        Results are reported in \autoref{tab:cars_abla}.
        Following the two previous ablations, we set $|\widetilde{\mathcal{B}}| = \ |\mathcal{B}| = 40$ and $N_v = \ N_t$.
        To avoid any bias in the selection of these parameters, we report results on the Cars-196 dataset for three different DML losses; namely the contrastive loss \cite{Chopra_CVPR_2005}, the triplet loss \cite{Schroff_2015_CVPR} and the binomial loss \cite{Ustinova_2016_NIPS}.
        We also compare three different approaches, namely: the \emph{Baseline}, the training class example generation only (denoted as TCG) and \ourmethod.
        For the three DML losses, both the training class example generation and \ourmethod \ lead to significant improvements over the \emph{Baseline}.
        \emph{E.g.}, with the contrastive loss, the \emph{Baseline} is improved from $74.0\%$ to $78.8\%$ which is nearly a $5\%$ improvement in Recall@1.
        Besides, the performances of the binomial loss and the triplet loss are improved from $71.2\%$ to $77.8\%$ and from $70.9\%$ to $73.6\%$ respectively, which is an absolute improvement of $+6.6\%$ and $+1.7\%$ in Recall@1.
        This improvements is achieved without tuning the parameters for the Cars-196 dataset and for all evaluated DML loss functions.

\setlength{\tabcolsep}{5pt}
\begin{table}[t!]
    \footnotesize
    \begin{center}
        \begin{tabular}{|c|cccc|}\hline
            Method & R@1 & R@2 & R@4 & R@8 \\\hline
            Contrastive & 74.0 & 83.1 & 89.4 & 93.8 \\
            Contrastive + TCG & \underline{76.3} & \underline{85.2} & \underline{90.8} & \underline{94.6} \\
            Contrastive + \ourmethod & \textbf{78.8} & \textbf{86.4} & \textbf{91.7} & \textbf{95.4} \\\hline
            Triplet& 70.9 & 80.5 & 87.6 & 92.8 \\
            Triplet + TCG & \underline{72.0} & \underline{81.4} & \underline{88.1} & \textbf{93.2}\\
            Triplet + \ourmethod & \textbf{73.6} & \textbf{82.2} & \textbf{88.5} & \textbf{93.2} \\
            \hline
            Binomial & 71.2 & 80.8 & 87.7 & 93.0 \\
            Binomial + TCG & \underline{74.6} & \underline{83.8} & \underline{89.7} & \underline{93.9} \\
            Binomial + \ourmethod & \textbf{77.8} & \textbf{86.1} & \textbf{91.3} & \textbf{94.7} \\
            \hline
        \end{tabular}
        \caption{Recall at $K$ for three loss functions on the Cars-196 dataset with GoogleNet backbone network. We compare the baseline, the training class example generation only (denoted TCG) and \ourmethod. Results that improve over the baseline are underlined and best results are in bold for each loss function.}
        \label{tab:cars_abla}
    \end{center}
\end{table}
        
\section{Comparison to the state-of-the-art}\label{sec:exp}
    \input{tex/sota_tables.tex}
    In this section, we present the benefits of \ourmethod \ on four deep metric learning datasets named Cub-200-2011 \cite{CUB_200_2011}, Cars-196 \cite{CARS_196}, Stanford Online Products \cite{Song_2016_CVPR} and In-Shop Clothes Retrieval \cite{Liu_2016_CVPR_INSHOP}.
    We follow the standard splits from \cite{Opitz_toap_PAMI} and Recall@K are reported for each dataset respectively in \autoref{tab:CUB}, \autoref{tab:CARS}, \autoref{tab:SOP} and \autoref{tab:SHOP}.
    
    We first compare our architecture with recent sample generation approaches from the literature using the now standard GoogleNet Backbone to ensure all results are fairly comparable.
    As we can see, \ourmethod \ obtains very strong results on all datasets.
    We achieve best performances on Cub-200-2011 and Cars-196, and second best on Stanford Online Products.
    This shows the importance of combining in class sample generation, like in \cite{Lin_2018_ECCV} with hard sample generation like \cite{Zheng_2019_CVPR}, which \ourmethod \ achieves with a simple architecture.
    
    In order to compare \ourmethod \ with recent methods, we also report Recall@K using BN-Inception \cite{Ioffe_2015_ICML} with the same hyper-parameters as the ones used for GoogleNet.
    \ourmethod \ obtains strong performances when compared to very recent state-of-the-art methods.
    On Cub-200-2011, we obtain second best performances, being only 0.4\% behind HORDE~\cite{Jacob_2019_ICCV}.
    On Cars-196 and Stanford Online Products, we obtain  performances comparable to that of Multi-Similarity loss \cite{Wang_2019_CVPR} and SoftTriplet \cite{Qian_2019_ICCV}.
    On In-Shop Clothes Retrieval, we obtain results comparable to MS loss \cite{Wang_2019_CVPR} and better than recently proposed D\&C \cite{Sanakoyeu_2019_CVPR} and MIC \cite{Roth_2019_ICCV} that use the stronger ResNet50 backbone network.
    We want to emphasize that the reported results were obtained using the constrastive loss function, and yet bring improvements to the baseline comparable to that of using a much more advance loss function such as \cite{Wang_2019_CVPR}, \cite{Qian_2019_ICCV} or \cite{Jacob_2019_ICCV}.
    We believe this demonstrates the soundness of our approach.

%% file: tex/sota_tables.tex
    \setlength{\tabcolsep}{1.5pt}

    \setlength{\tabcolsep}{5pt}
    \begin{table}
        \centering
        \begin{tabular}{|c|c|c|c|c|c|}\hline
            Backbone & Method & R@1 & R@2 & R@4 & R@8 \\\hline
            \multirow{5}{*}{GoogleNet} & DAMLRMM \cite{Xu_2019_CVPR} & 55.1 & 66.5 & 76.8 & 85.3 \\
             & DAML \cite{Duan_2018_CVPR} & 52.7 & 65.4 & 75.5 & 84.3 \\
             & DVML \cite{Lin_2018_ECCV} & 52.7 & 65.1 & 75.5 & 84.3  \\
             & HDML \cite{Zheng_2019_CVPR} & 53.7 & 65.7 & 76.7 & 85.7 \\
             \cline{2-6}
             & \textbf{\ourmethod \ (Ours)} & \textbf{59.7} & \textbf{71.1} & \textbf{80.4} & \textbf{88.1} \\\hline
             \multirow{4}{*}{BN-Inception} & MS loss \cite{Wang_2019_CVPR} & 65.7 & 77.0 & \textbf{86.3} & \textbf{91.2} \\
             & SoftTriplet \cite{Qian_2019_ICCV} & 65.4 & 76.4 & 84.5 & 90.4 \\
             & HORDE \cite{Jacob_2019_ICCV} & \textbf{66.8} & \underline{77.4} & \underline{85.1} & \underline{91.0} \\
             \cline{2-6}
             & \textbf{\ourmethod \ (Ours)} & \underline{66.4} & \textbf{78.9} & 84.6 & 90.3 \\\hline
        \end{tabular}
        \caption{Comparison to the state-of-the-art on the Cub-200-2011 dataset. Results are reported using GoogleNet as backbone network for fair comparison with generation-based methods. Results are also reported with BN-Inception backbone for comparison with other recent methods.}
        \label{tab:CUB}
    \end{table}

    \begin{table}
        \centering
        \begin{tabular}{|c|c|c|c|c|c|}\hline
            Backbone & Method & R@1 & R@2 & R@4 & R@8 \\\hline
            \multirow{5}{*}{GoogleNet} & DAMLRMM \cite{Xu_2019_CVPR} & 73.5 & 82.6 & 89.1 & 93.5 \\
             & DAML \cite{Duan_2018_CVPR}  & 75.1 & 83.8 & 89.7 & 93.5 \\
             & DVML \cite{Lin_2018_ECCV} & \underline{82.0} & \underline{88.4} & \underline{93.3} & \textbf{96.3} \\
             & HDML \cite{Zheng_2019_CVPR} & 79.1 & 87.1 & 92.1 & 95.5 \\
             \cline{2-6}
             & \textbf{\ourmethod \ (Ours)} & \textbf{82.1} & \textbf{89.1} & \textbf{93.6} & \underline{96.2} \\\hline
             \multirow{4}{*}{BN-Inception} & MS loss \cite{Wang_2019_CVPR} & 84.1 & 90.4 & 94.0 & 96.5 \\
             & SoftTriplet \cite{Qian_2019_ICCV} & 84.5 & 90.7 & 94.5 & 96.9 \\
             & HORDE \cite{Jacob_2019_ICCV} & \textbf{86.2} & \textbf{91.9} & \textbf{95.1} & \textbf{97.2} \\
             \cline{2-6}
             & \textbf{\ourmethod \ (Ours)} & 83.9 & 90.3 & 94.4 & 96.9 \\\hline
        \end{tabular}
        \caption{Comparison to the state-of-the-art on the Cars-196 dataset. Results are reported using GoogleNet as backbone network for fair comparison with generation-based methods. Results are also reported with BN-Inception backbone for comparison with other recent methods.}
        \label{tab:CARS}
    \end{table}
    
    \begin{table}
        \centering
        \begin{tabular}{|c|c|c|c|c|}\hline
            Backbone & Method & R@1 & R@10 & R@100 \\\hline
            \multirow{5}{*}{GoogleNet} & DAMLRMM \cite{Xu_2019_CVPR} & 69.7 & \textbf{85.2} & \underline{93.2} \\
             & DAML \cite{Duan_2018_CVPR} & 68.4 & 83.5 & 92.3 \\
             & DVML \cite{Lin_2018_ECCV} & \textbf{70.2} & \textbf{85.2} & \textbf{93.8} \\
             & HDML \cite{Zheng_2019_CVPR} & 68.7 & 83.2 & 92.4 \\
             \cline{2-5}
             & \textbf{\ourmethod \ (Ours)} & \underline{69.9} & 84.3 & 92.6 \\\hline
            \multirow{2}{*}{ResNet50} & D\&C \cite{Sanakoyeu_2019_CVPR} & 75.9 & 88.4 & 94.9 \\
             & MIC \cite{Roth_2019_ICCV} & 77.2 & 89.4 & 95.6 \\\hline
             \multirow{4}{*}{BN-Inception} & MS loss \cite{Wang_2019_CVPR} & 78.2 & 90.5 & 96.0 \\
             & SoftTriplet \cite{Qian_2019_ICCV} & \underline{78.3} &  90.3 & 95.9 \\
             & HORDE \cite{Jacob_2019_ICCV} & \textbf{80.1} & \textbf{91.3} & \underline{96.2} \\
             \cline{2-5}
             & \textbf{\ourmethod \ (Ours)} & 76.0 & 88.3 & 94.8 \\\hline
        \end{tabular}
        \caption{Comparison to the state-of-the-art on the Stanford Online Products dataset. Results are reported using GoogleNet as backbone network for fair comparison with generation-based methods. Results are also reported with BN-Inception backbone for comparison with other recent methods.}
        \label{tab:SOP}
    \end{table}
    
    \begin{table}
        \centering
        \begin{tabular}{|c|c|c|c|c|c|}\hline
            Backbone & Method & R@1 & R@10 & R@20 & R@30 \\\hline
            \multirow{2}{*}{ResNet50} & D\&C \cite{Sanakoyeu_2019_CVPR} & 85.7 & 95.5 & 96.9 & 97.5 \\
             & MIC \cite{Roth_2019_ICCV} & 88.2 & 97.0 & - & 98.0 \\\hline
            \multirow{3}{*}{BN-Inception} & MS loss \cite{Wang_2019_CVPR} & \underline{89.7} & \textbf{97.9} & \textbf{98.5} & \textbf{98.8} \\
            & HORDE \cite{Jacob_2019_ICCV} & \textbf{90.4} & \underline{97.8} & \underline{98.4} & \underline{98.7} \\
             \cline{2-6}
            & \textbf{\ourmethod \ (Ours)} & 89.1 & 97.0 & 97.9 & 98.3 \\\hline
        \end{tabular}
        \caption{Comparison to the state-of-the-art on the In-Shop Clothes Retrieval dataset.}
        \label{tab:SHOP}
    \end{table}

%% file: tex/ccl.tex
\section{Conclusion}\label{sec:ccl}
    In this paper, we introduce \ourmethod, a generation-based strategy that naturally solves the generation of hard examples.
    \ourmethod \ naturally solves the problem of generating incorrectly labeled hard negative examples by relying on a set of virtual class prototypes solely composed of generated examples.
    Even when the generator produces examples beyond their class manifolds, the presence of virtual classes ensures that the examples are still generated with the correct labels regarding the training classes.
    We empirically show that \ourmethod \ outperforms the state-of-the-art mining strategies and leads to competitive results when compared to complementary approaches.  
    This is validated on four deep metric learning datasets named Cub-200-2011, Cars-196, Stanford Online Products and In-Shop Clothes Retrieval.

%% file: main.bbl
\begin{thebibliography}{10}
\providecommand{\url}[1]{\texttt{#1}}
\providecommand{\urlprefix}{URL }
\providecommand{\doi}[1]{https://doi.org/#1}

\bibitem{Carvalho_2018_SIGIR}
Carvalho, M., Cad\`{e}ne, R., Picard, D., Soulier, L., Thome, N., Cord, M.:
  Cross-modal retrieval in the cooking context: Learning semantic text-image
  embeddings. In: The 41st International ACM SIGIR Conference on Research \&
  Development in Information Retrieval (2018)

\bibitem{Chen_2017_CVPR}
Chen, W., Chen, X., Zhang, J., Huang, K.: Beyond triplet loss: A deep
  quadruplet network for person re-identification. In: The IEEE Conference on
  Computer Vision and Pattern Recognition (CVPR) (July 2017)

\bibitem{Chopra_CVPR_2005}
Chopra, S., Hadsell, R., LeCun, Y.: Learning a similarity metric
  discriminatively, with application to face verification. In: The IEEE
  Conference on Computer Vision and Pattern Recognition (CVPR) (2005)

\bibitem{Duan_2018_CVPR}
Duan, Y., Zheng, W., Lin, X., Lu, J., Zhou, J.: Deep adversarial metric
  learning. In: The IEEE Conference on Computer Vision and Pattern Recognition
  (CVPR) (June 2018)

\bibitem{Harwood_2017_ICCV}
Harwood, B., Kumar B~G, V., Carneiro, G., Reid, I., Drummond, T.: Smart mining
  for deep metric learning. In: The IEEE International Conference on Computer
  Vision (ICCV) (Oct 2017)

\bibitem{Ioffe_2015_ICML}
Ioffe, S., Szegedy, C.: Batch normalization: Accelerating deep network training
  by reducing internal covariate shift. In: Proceedings of the 32nd
  International Conference on Machine Learning (Jul 2015)

\bibitem{Jacob_2019_ICCV}
Jacob, P., Picard, D., Histace, A., Klein, E.: Metric learning with horde:
  High-order regularizer for deep embeddings. In: The IEEE International
  Conference on Computer Vision (ICCV) (October 2019)

\bibitem{Kim_2018_ECCV}
Kim, W., Goyal, B., Chawla, K., Lee, J., Kwon, K.: Attention-based ensemble for
  deep metric learning. In: The European Conference on Computer Vision (ECCV)
  (September 2018)

\bibitem{CARS_196}
Krause, J., Stark, M., Deng, J., Fei-Fei, L.: 3d object representations for
  fine-grained categorization. In: 4th International IEEE Workshop on 3D
  Representation and Recognition (3dRR-13) (Dec 2013)

\bibitem{Lin_2018_ECCV}
Lin, X., Duan, Y., Dong, Q., Lu, J., Zhou, J.: Deep variational metric
  learning. In: The European Conference on Computer Vision (ECCV) (September
  2018)

\bibitem{Liu_2016_CVPR}
Liu, H., Tian, Y., Yang, Y., Pang, L., Huang, T.: Deep relative distance
  learning: Tell the difference between similar vehicles. In: The IEEE
  Conference on Computer Vision and Pattern Recognition (CVPR) (June 2016)

\bibitem{Liu_2016_CVPR_INSHOP}
Liu, Z., Luo, P., Qiu, S., Wang, X., Tang, X.: Deepfashion: Powering robust
  clothes recognition and retrieval with rich annotations. In: Proceedings of
  IEEE Conference on Computer Vision and Pattern Recognition (CVPR) (June 2016)

\bibitem{Movshovitz-Attias_2017_ICCV}
Movshovitz-Attias, Y., Toshev, A., Leung, T.K., Ioffe, S., Singh, S.: No fuss
  distance metric learning using proxies. In: The IEEE International Conference
  on Computer Vision (ICCV) (Oct 2017)

\bibitem{Song_2017_CVPR}
Oh~Song, H., Jegelka, S., Rathod, V., Murphy, K.: Deep metric learning via
  facility location. In: The IEEE Conference on Computer Vision and Pattern
  Recognition (CVPR) (July 2017)

\bibitem{Song_2016_CVPR}
Oh~Song, H., Xiang, Y., Jegelka, S., Savarese, S.: Deep metric learning via
  lifted structured feature embedding. In: The IEEE Conference on Computer
  Vision and Pattern Recognition (CVPR) (June 2016)

\bibitem{Opitz_2017_ICCV}
Opitz, M., Waltner, G., Possegger, H., Bischof, H.: Bier - boosting independent
  embeddings robustly. In: The IEEE International Conference on Computer Vision
  (ICCV) (Oct 2017)

\bibitem{Opitz_toap_PAMI}
Opitz, M., Waltner, G., Possegger, H., Bischof, H.: Deep metric learning with
  {BIER:} boosting independent embeddings robustly. IEEE transactions on
  pattern analysis and machine intelligence  (2018)

\bibitem{Qian_2019_ICCV}
Qian, Q., Shang, L., Sun, B., Hu, J., Li, H., Jin, R.: Softtriple loss: Deep
  metric learning without triplet sampling. In: The IEEE International
  Conference on Computer Vision (ICCV) (October 2019)

\bibitem{Rippel_2016_ICLR}
Rippel, O., Paluri, M., Dollar, P., Bourdev, L.: Metric learning with adaptive
  density discrimination. International Conference on Learning Representations
  (ICLR)  (May 2016)

\bibitem{Roth_2019_ICCV}
Roth, K., Brattoli, B., Ommer, B.: Mic: Mining interclass characteristics for
  improved metric learning. In: The IEEE International Conference on Computer
  Vision (ICCV) (October 2019)

\bibitem{Sanakoyeu_2019_CVPR}
Sanakoyeu, A., Tschernezki, V., Buchler, U., Ommer, B.: Divide and conquer the
  embedding space for metric learning. In: The IEEE Conference on Computer
  Vision and Pattern Recognition (CVPR) (June 2019)

\bibitem{Schroff_2015_CVPR}
Schroff, F., Kalenichenko, D., Philbin, J.: Facenet: A unified embedding for
  face recognition and clustering. In: The IEEE Conference on Computer Vision
  and Pattern Recognition (CVPR) (June 2015)

\bibitem{Sohn_2016_NIPS}
Sohn, K.: Improved deep metric learning with multi-class n-pair loss objective.
  In: Advances in Neural Information Processing Systems 29 (Dec 2016)

\bibitem{Szegedy_2015_CVPR}
Szegedy, C., Liu, W., Jia, Y., Sermanet, P., Reed, S., Anguelov, D., Erhan, D.,
  Vanhoucke, V., Rabinovich, A.: Going deeper with convolutions. In: IEEE
  Conference on Computer Vision and Pattern Recognition (CVPR) (June 2015)

\bibitem{Ustinova_2016_NIPS}
Ustinova, E., Lempitsky, V.: Learning deep embeddings with histogram loss. In:
  Advances in Neural Information Processing Systems 29 (Dec 2016)

\bibitem{CUB_200_2011}
Wah, C., Branson, S., Welinder, P., Perona, P., Belongie, S.: {The Caltech-UCSD
  Birds-200-2011 Dataset}. Tech. Rep. CNS-TR-2011-001, California Institute of
  Technology (2011)

\bibitem{Wang_2018_CVPR}
Wang, H., Wang, Y., Zhou, Z., Ji, X., Gong, D., Zhou, J., Li, Z., Liu, W.:
  Cosface: Large margin cosine loss for deep face recognition. In: The IEEE
  Conference on Computer Vision and Pattern Recognition (CVPR) (June 2018)

\bibitem{Wang_2017_ICCV}
Wang, J., Zhou, F., Wen, S., Liu, X., Lin, Y.: Deep metric learning with
  angular loss. In: The IEEE International Conference on Computer Vision (ICCV)
  (Oct 2017)

\bibitem{Wang_2019_CVPR}
Wang, X., Han, X., Huang, W., Dong, D., Scott, M.R.: Multi-similarity loss with
  general pair weighting for deep metric learning. In: The IEEE Conference on
  Computer Vision and Pattern Recognition (CVPR) (June 2019)

\bibitem{Wehrmann_2018_CVPR}
Wehrmann, J., Barros, R.C.: Bidirectional retrieval made simple. In: The IEEE
  Conference on Computer Vision and Pattern Recognition (CVPR) (June 2018)

\bibitem{Xu_2019_CVPR}
Xu, X., Yang, Y., Deng, C., Zheng, F.: Deep asymmetric metric learning via rich
  relationship mining. In: The IEEE Conference on Computer Vision and Pattern
  Recognition (CVPR). pp. 4076 -- 4085 (June 2019)

\bibitem{Xuan_2018_ECCV}
Xuan, H., Souvenir, R., Pless, R.: Deep randomized ensembles for metric
  learning. In: The European Conference on Computer Vision (ECCV) (September
  2018)

\bibitem{Yu_2018_ECCV}
Yu, B., Liu, T., Gong, M., Ding, C., Tao, D.: Correcting the triplet selection
  bias for triplet loss. In: The European Conference on Computer Vision (ECCV)
  (September 2018)

\bibitem{Yuan_2017_ICCV}
Yuan, Y., Yang, K., Zhang, C.: Hard-aware deeply cascaded embedding. In: The
  IEEE International Conference on Computer Vision (ICCV) (Oct 2017)

\bibitem{Zhao_2018_ECCV}
Zhao, Y., Jin, Z., Qi, G.j., Lu, H., Hua, X.s.: An adversarial approach to hard
  triplet generation. In: The European Conference on Computer Vision (ECCV)
  (September 2018)

\bibitem{Zheng_2019_CVPR}
Zheng, W., Chen, Z., Lu, J., Zhou, J.: Hardness-aware deep metric learning. In:
  The IEEE Conference on Computer Vision and Pattern Recognition (CVPR) (June
  2019)

\end{thebibliography}
